\documentclass[12pt]{article}

\usepackage{sbc-template}

\usepackage{graphicx,url}

\usepackage[T1]{fontenc} % Codificação para fontes com suporte a caracteres acentuados
\usepackage[utf8]{inputenc} % Codificação UTF-8 (necessário apenas para pdflatex)
\usepackage{multicol}
\usepackage{multirow}
\usepackage{graphicx,url}
\usepackage{amsmath,amssymb,amsfonts}
\usepackage{algorithmic}
\usepackage{textcomp}
\usepackage{url}
\usepackage{enumerate}
\usepackage{multirow}
\usepackage{soul}
\usepackage{verbatim}
\usepackage[table,xcdraw]{xcolor}
\usepackage[shortlabels]{enumitem}
\usepackage{scalefnt} % pacote para diminuir fonte das tabelas
\usepackage{comment} % pacote de comentarios
\usepackage{lscape}
\usepackage{fancyhdr}
\usepackage{sbgames}
\usepackage[super]{nth}
\usepackage{hyperref}
\usepackage{booktabs}

\emergencystretch=\maxdimen
\trilha{Computação} %Preencha com: Artes \& Design | Computação | Cultura | Saúde | Indústria | Educação | Tutoriais | MAGICA | WGP-Jogos
\local{Salvador/BA} % Exemplo: Manaus/AM
\ano{2025} % Exemplo: 2024
\edicao{XXIV} %Exemplo: XXIII

\title{Boardwalk: Towards a Framework for Creating Board Games with LLMs}

\author{Álvaro Guglielmin Becker\inst{1}, Gabriel Bauer de Oliveira\inst{1},\\ Lana Bertoldo Rossato\inst{1}, Anderson Rocha Tavares\inst{1} }

\address{Instituto de Informática -- Universidade Federal do Rio Grande do Sul (UFRGS)\\
  Caixa Postal 15064 -- Porto Alegre -- RS -- Brasil
  \email{\{agbecker,gabriel.bauer,lbrossato,artavares\}@inf.ufrgs.br}
}

\begin{document} 

\maketitle

\thispagestyle{plain}

\begin{abstract}
\textbf{Introduction}: Implementing board games in code can be a time-consuming task. However, Large Language Models (LLMs) have been proven effective at generating code for domain-specific tasks with simple contextual information.
\textbf{Objective:} We aim to investigate whether LLMs can implement digital versions of board games from rules described in natural language. This would be a step towards an LLM-assisted framework for quick board game code generation. We expect to determine the main challenges for LLMs to implement the board games, and how different approaches and models compare to one another.
\textbf{Methodology}: We task three state-of-the-art LLMs (Claude, DeepSeek and ChatGPT) with coding a selection of 12 popular and obscure games in free-form and within Boardwalk, our proposed General Game Playing API. We anonymize the games and components to avoid evoking pre-trained LLM knowledge. The implementations are tested for playability and rule compliance. We evaluate success rate and common errors across LLMs and game popularity.
\textbf{Results:} Our approach proves viable, with the best performing model, Claude 3.7 Sonnet, yielding 55.6\% of games without any errors. While compliance with the API increases error frequency, the severity of errors is more significantly dependent on the LLM. We outline future steps for creating a framework to integrate this process, making the elaboration of board games more accessible.
\end{abstract}


\section{Introduction}
Board games are an important object of study in the field of Artificial Intelligence (AI), with many works concerning themselves with developing algorithms for playing or modifying games \cite{Browne2011evolutionary,todd2024gavel}. However, in order for such studies to be conducted, they first require the researchers to represent the game they're studying in code. This can be a time-consuming task, especially for games of higher complexity with various different pieces and edge cases to consider.

For this reason, several Game Description Languages (GDLs), such as Ludii \cite{ludii}, have been developed; these are domain-specific languages meant to facilitate the implementation and analysis of digital board games, by providing direct commands for defining the components and rules of such games. However, creating game descriptions with GDLs still requires knowledge of the language's own syntax, which sometimes requires convoluted constructions for simple operations of specific games\footnote{For example, Ultimate Tic-Tac-Toe in Ludii's GDL has 9 levels of nested tests to check move validity \cite{ultimateTictactoe}.}. %  and taking time to program the desired game with it.

Therefore, Large Language Models (LLMs) present themselves as highly appealing tools to aid in this task. Modern LLMs have shown great performance in code generation tasks \cite{Coignion2024llmcoding}, adding significant efficiency to code-dependent workflows. One particularly interesting aspect of LLMs is their capacity for In-Context Learning (ICL), being able to perform tasks they had not been trained for simply from receiving additional information pertaining to those tasks as part of the prompt \cite{few-shot-learners}.

In this work, we aim to answer the following research question: \textbf{can LLMs implement digital versions of board games from rules described in natural language?} To answer the question, we ask state-of-the-art LLMs (DeepSeekV3, Claude 3.7 Sonnet, and ChatGPT-4o) to code a selection of 12 games in Python, both in free-form and within Boardwalk, a standardized API we provide. We also request the LLMs to adapt their free-form implementations to the API. The games and their components are anonymized to evaluate the LLMs' understanding of rules in natural language rather than evoking code associated with such names in pre-training data. Furthermore, 6 of the 12 games are rather obscure, such that it is even less likely that the LLMs evoke pre-training code associated with their rules.
The resulting game implementations are tested for playability and rule compliance. We enumerate common errors across games and LLMs to better understand their relations.

Our main findings are that LLMs are largely capable of implementing games from rule descriptions, both free-form and using the API, and that Claude is the best-performing of the evaluated models.

Our main contributions can be summarized as follows:
\begin{itemize}
    \item A broad evaluation of LLM capabilities on generating digital versions of board games from rules in natural language;
    \item Boardwalk, a standardized API for General Game Playing, easing the integration of game implementations and game-playing AIs. The API aims for simplicity by allowing the game definition in Python rather than a Game Description Language with its own learning curve. Boardwalk code is open-sourced at \url{https://github.com/LabCRAIG/boardwalk}.
\end{itemize}

The remainder of this paper is organized as follows: Section \ref{sec:related} presents related work. Section \ref{sec:boardwalk} introduces the Boardwalk API. Section \ref{sec:methodology} details our evaluation methodology. % rules and prompting the LLMs to generate code from them, both free-form and in accordance with the API. 
Section \ref{sec:results} presents quantitative results of the error rates of each model and game, as well as a qualitative analysis of specific errors and recurring challenges observed. Finally, Section \ref{sec:conclusion} summarizes our findings and outlines possible future work towards an AI-powered board game generation framework. % integrating the process described here.
%explore the possibility of developing a framework for generating playable code for board games based on natural language descriptions of their rules. We contribute a Python API for creating board games with a standardized interface for both human and AI players, and verify the capacity of three state-of-the-art language models to create code both with and without the API, from descriptions of a diverse set of games. We compare the performance of the different models and the pros and cons of different approaches to code generation, and analyze possibilities to eventually develop a low-complexity yet dependable framework for automated digital board game generation.

\section{Related Work}
\label{sec:related}
% Most existing studies use the Ludii game description language \cite{ludii}.

Most existing studies in LLM-oriented procedural content generation for games are centered on video games. \cite{todd2023levels} were able to train LLMs to create novel and functional Sokoban levels. \cite{mariogpt} were able to fine-tune GPT2 to create Super Mario Bros levels with custom features defined by natural language prompts. LLMs have also been used to tell interactive narratives in games \cite{Sun2023, Kumaran2023}. All of these focused on creating content for otherwise human-implemented games.

Using a video game description language, \cite{hu2024} had LLMs generate both rules and levels for maze-based games. They found that the models required context-rich prompts for optimal performance and faced greater challenges with the logical aspects of game design compared to the syntax of the language.

\cite{Browne2011evolutionary} reports the seemingly first automated system to generate a commercial board game from scratch\footnote{Yavalath: \url{https://boardgamegeek.com/boardgame/33767/yavalath}}. It uses evolutionary algorithms to evolve game rules on the Ludi (with one `i') Game Description Language (GDL), combined with automated playtesting with general game playing (GGP) \cite{Genesereth2005ggp} algorithms. However, the Ludi framework is not available for further experimentation. 

The second version of Ludii (now with two `i') GDL \cite{ludii}, had an initial focus on game archaeology and not on automated game creation. However, recent work has managed to use Ludii in automated board game creation. For example,  \cite{todd2024gavel} fine-tuned an LLM in the Ludii GDL. Using a fill-in-the-middle approach, the authors removed parts of pre-existing game descriptions and tasked the LLM with completing them, generating potentially different rules, without having to create new games from scratch. However, generated rules often contained introns, i.e. elements that were never used in gameplay.

\cite{ludii-grammar-llm} used ICL to generate Ludii implementations from natural language rule descriptions. However, due to the complexity of the Ludii grammar, they followed a process for each game of iteratively finding a minimal valid subgrammar to describe the game, and then iteratively finding a description over that grammar that correctly implements the rules. Not only was this very computationally demanding, requiring repeated inferences for each game, but it further showed that using the Ludii language requires a significant amount of upfront work to familiarize the LLM with its structure. Only 72\% of descriptions generated by their method compiled (i.e. were valid in the Ludii GDL grammar).

%We leverage the strengths of \cite{todd2024gavel,ludii-grammar-llm} by using a workflow based on an API with simple documentation, requiring the model to simply implement a few methods for the desired game. However, g
Given the results of \cite{todd2024gavel,ludii-grammar-llm}, we decided not to use a specific GDL for our purposes, due to the significant limitations in code generation that stem from the LLMs' lack of familiarity with them. Instead, our Boardwalk GGP API (see Section \ref{sec:boardwalk}) aims for simplicity: the games are defined as Python programs. Using a programming language is easier for LLMs as they consume extensive programming content in their pre-training \cite{Jiang2024llmcode}. Thus, the LLMs are likely to have a significantly higher success rate writing code in a popular programming language rather than a specific GDL. Moreover, this removes an entry barrier for human programmers as well, because there is no extra learning curve for a new language.

\section{The Boardwalk API}
\label{sec:boardwalk}
The Boardwalk API is a General Game Playing (GGP) platform, with the purpose of acting as a tool for scientific study. It provides an interface for defining a game and having it played via a standardized input and output format. Boardwalk was planned to minimize the amount of code needed to implement a game: boilerplate code (user interface, game state, and game loop) is handled by Boardwalk. This requires the programmer to implement only the logic specific to the game rules in Python, essentially providing a game engine that is easy to work with for both human programmers and LLMs.

For the scope of this study, the API's capacity for representation was limited to perfect information games with any number of players whose boards can be modeled in a rectangular grid, with a simple command-line user interface. However, it would be reasonably simple to expand its breadth of representable games, as well as to add a graphical user interface, without breaking any code developed in the current stage.

The code and its documentation are available on GitHub, with the release \textit{Boardwalk 1.0} corresponding to the version used in this study\footnote{\url{https://github.com/LabCRAIG/boardwalk/releases/tag/rules-phase-no-ai}}. That version does not support game-playing AI agents. This feature has been added to the project's main branch.  

Next, we present an overview of how the API is structured, what resources it offers to the programmer, and what needs to be implemented for a given game.

%The API uses a top-down approach; individual components carry no logic on their own, with 
The implementation of the game rules is divided into the validity of moves, the order of play, and win conditions. %Pieces and auxiliary variables serve only as values to be checked in that paradigm.
Two classes are specified: \textbf{Board} and \textbf{Game}.

The \textbf{Board} class is final (non-extensible), and implements the matrix holding the board layout (spaces and pieces), as well as methods to change piece positioning by placement, moving, or removal. 
The \textbf{Game} class is designed to hold the logic of the game's rules. By default, it implements a series of attributes and methods responsible for holding and representing the game state, interfacing with the player, and operating the standard game loop. However, the class itself cannot be used, as it has several vital methods left blank. Hence, it must be inherited into a child class specific to the game being implemented, and the blank methods are overridden there, implementing the game rules.

At the bare minimum, the programmer needs only to define a custom enumeration to represent the game's players and override four methods of the Game class:

\begin{itemize}
    \item \texttt{validate\_move}: receives a move and returns a boolean indicating whether the move is valid according to the game rules.
    \item \texttt{game\_finished}: receives a game state and returns a boolean indicating if, according to the rules, the  game state is terminal. This is where the win conditions are implemented.
    \item \texttt{get\_winner}: returns an integer corresponding to the winning player (according to the defined enumeration). In case of a draw, it instead returns None.
    \item \texttt{next\_player}: returns an integer corresponding to the player to make the next move (according to the defined enumeration).
\end{itemize}

These four methods have no preexisting definition, and thus must be overridden for all games (obligatory methods). None of them alter the game state; provided the return values are of the expected types, the API handles the control flow and state of the game.

The other Game class methods all have existing default implementations. Thus, they may suffice for simpler games, given the obligatory methods are implemented. However, they may also be overridden to implement more complex rules or customize the user interface. The one most often modified will be the \texttt{perform\_move} method, which is responsible for altering the game board according to the informed move once validated. Programmers may opt to add other modifications to the board as part of this method to allow for side effects of moves (such as captures and promotions in Checkers, for example).

The Game class also has three attributes, which are set during initialization and all child classes must account for: \texttt{board}, which is an instance of the Board class; \texttt{round}, an integer which counts the number of turns played, and by default increases by one with every move; and \texttt{current\_player}, an integer specifying the programmer-defined enumeration of the next player to make a move. These attributes define the game state.

All subclasses of the Game class follow the same game loop: for every turn, while the game is not finished, the current state of the board is printed. The current player is repeatedly prompted to make a move, until their input is deemed valid by \texttt{validate\_move}. The move is then performed by \texttt{perform\_move}, along with its side effects. The game then checks if a terminal condition has been met. If yes, the winner is identified and printed, and the game loop ends. If not, the \texttt{round} counter is increased and the next player is assigned, and the loop continues.

\section{Methodology}
\label{sec:methodology}
This section describes the methods used to answer our research question, i.e., whether LLMs can implement digital versions of board games from rules described in natural language. We conducted tests on three different, state-of-the-art LLMs: DeepSeekV3 \cite{deepseekai2025deepseekv3technicalreport}, Claude 3.7 Sonnet \cite{claude3}, and ChatGPT-4o \cite{openai2024gpt4technicalreport}. These LLMs were accessed via the Poe website\footnote{\url{https://poe.com/}}. They will hereby be referred to as DeepSeek, Claude and GPT, respectively.

We selected 12 board games. We considered 6 of these (Tic-Tac-Toe, Peg Solitaire, Reversi, Nine Men's Morris, Checkers and Chess) to be widely popular games, and the remaining 6 (Ard Ri, Domineering, Tron, Amazons, Kharebga and Unashogi) to be more obscure games\footnote{Game descriptions can be found in the Ludii database \url{https://ludii.games/library.php}}. We considered it likely that the LLMs would be significantly more acquainted with the rules of the well-known games and have implementations of them in code as part of their training data, which would make it easier for them to implement these games. As such, the obscure games might offer a better measure of the models' capacity for generalization for this task---understanding the game rules and implementing code based purely on natural language descriptions without evoking pre-existing game-specific knowledge.

An additional step taken to avoid evoking pre-trained code associated with the games was their anonymization. We removed as much identifying information as possible: each game was given a fake name, and pieces received unusual names. For example, Tic-Tac-Toe was named ``Peridot'' and the pieces were A and V instead of X and O. Moreover, all game rules were described in a standardized structure. 

Each experiment consists of one model implementing one game. For each experiment, a new chat was initiated, and a series of standardized prompts with information regarding the game were given to the model, finally asking it to produce Python code to implement it. The prompts used are available on GitHub\footnote{\url{https://github.com/LabCRAIG/boardwalk-prompts}}.

To gain a better understanding of LLMs' capabilities and how much the API impacted their performance, three sets of tests were performed, each presenting the model with different context information (we refer to these as code-generation modes):

\begin{enumerate}
    \item the Boardwalk API (Section \ref{sec:boardwalk}) documentation and the game rules (\textit{API implementation});
    \item only the game rules (\textit{independent}, free-form, implementation);
    \item the API documentation, the game rules, and the independent implementation of the same game by the same model (\textit{adapted} implementation, where the LLM had to adapt the \textit{independent} code to be API-compliant).
\end{enumerate}

We have shown only the documentation of the API to the models, not its code. In total, 108 experiments were run (12 games, 3 models, 3 code-generation modes), with each yielding Python code for one game.

The outputs were evaluated by manually running the generated code, checking the initial board setup, and manually playing one or more standard rounds of the game to verify if the rules were correctly implemented. In cases where the code could not be run due to a syntax error or crash, a manual inspection of the code was made in an attempt to identify the cause. If the issue was easily spotted and fixable, an altered version of the code with that issue corrected was run, to evaluate the game logic despite those errors (though they were still registered). Examples of these fixable issues are given in Section~\ref{sec:qualitative}.

\section{Results}
\label{sec:results}
In all experiments, all of the models output Python code as requested. All of the results were code for an interactive game, save for one: DeepSeek's independent implementation of Amazons did not take inputs from a player or run a game loop; it simply printed a sequence of sample moves in the game, demonstrating how it worked. 

The results are divided into quantitative analysis (Section \ref{sec:quantitative}), where we present the success and error rates of each model, and qualitative analysis (Section \ref{sec:qualitative}), where we examine specific types of errors and their manifestation and severity across models and games.

The generated code for all tests is available on GitHub\footnote{\url{https://github.com/LabCRAIG/boardwalk-prompts/tree/main/Results}}.

\subsection{Quantitative Analysis}
\label{sec:quantitative}

The generated codes were evaluated for the presence of the following types of errors: Python error (programming errors preventing the game from running); API (e.g. not implementing a mandatory method, or using the input/output API standard incorrectly); move verification (e.g. invalid moves allowed, valid moves not allowed); ending (game ends improperly, winner incorrectly identified); additional effect (e.g. captures and promotions not resolved properly); board error (e.g. incorrect layout or initial setup) and turn order (e.g. players are improperly skipped or called to move when they shouldn't).

Error occurrence was treated as a binary, i.e. we did not count multiple occurrences of the same error type, as our focus was to measure the occurrence of different error types. This allows us to gain insight into what aspects are more challenging for the LLMs, as well as to compare how the different models performed on the same task. A more detailed analysis of specific errors is given in the \textit{Qualitative Analysis} subsection.

Tables \ref{tab:combined-errors} show the occurrences of each type of error across each experiment group, for Claude, DeepSeek, and GPT, respectively. We remark that each LLM executed a total of 36 implementations (12 games, 3 code-generation modes)\footnote{As DeepSeek's independent implementation of Amazons was far outside the expected format, we simply counted it as a Python error, without checking for other error types.}. The total number of errors may exceed 36, as the same implementation can have multiple types of errors. 
Claude and DeepSeek showed worse results when using the Boardwalk API compared to the independent implementations, the latter accounting for only 16.7\% and 21.9\% of the total errors for each model, respectively. For GPT, it is only the adapted implementations that show a more significant increase in error frequency, being responsible for 43.6\% of total errors. 

\begin{table}[ht]
    \centering
    \caption{Distribution of error types by model and code-generation mode (API-compliant, Indep. for independent/free-form and Adap. for adapting from independent to API). The \textbf{Total (\%)} row shows the proportion of errors for each code-generation mode within each model, calculated as the count in each column divided by the total number of errors for that model. }
    \begin{tabular}{lccc | ccc | ccc}
        \toprule
         
        & \multicolumn{3}{c}{\textbf{Claude}} 
        & \multicolumn{3}{|c|}{\textbf{DeepSeek}} 
        & \multicolumn{3}{c}{\textbf{GPT}} \\
        \cmidrule(lr){2-4} \cmidrule(lr){5-7} \cmidrule(lr){8-10}
        \textbf{Error Type} & API & Indep. & Adap. 
        & API & Indep. & Adap. 
        & API & Indep. & Adap. \\
        \midrule
        Python      & 0 & 1 & 1 & 0 & 3 & 4 & 0 & 2 & 6 \\
        API         & 1 & - & 2 & 0 & - & 3 & 3 & - & 3 \\
        Move        & 4 & 2 & 4 & 6 & 1 & 4 & 4 & 4 & 5 \\
        Ending      & 5 & 1 & 1 & 4 & 2 & 3 & 5 & 4 & 4 \\
        Effect      & 2 & 1 & 3 & 4 & 3 & 2 & 1 & 2 & 3 \\
        Board       & 3 & 1 & 1 & 1 & 2 & 2 & 1 & 2 & 2 \\
        Turn order  & 1 & 0 & 2 & 3 & 0 & 1 & 1 & 2 & 1 \\
        \midrule
        \textbf{Total} 
                    & 16 & 6 & 14 & 18 & 11 & 19 & 15 & 16 & 24 \\
        %\textbf{Total (\%)} & 44.4\% & 16.7\% & 38.9\% & 38.3\% & 21.3\% & 40.4\% & 27.3\% & 29.1\% & 43.6\% \\
        \textbf{Total (\%)} & 44.4 & 16.7 & 38.9 & 37.5 & 22.9 & 39.6 & 27.3 & 29.1 & 43.6 \\
        \bottomrule
    \end{tabular}
    \label{tab:combined-errors}
\end{table}

Table \ref{tab:success-failure} shows the number of games each model was able to implement perfectly, as well as the number of unplayable games. The first is an important measure of success, whereas the second indicates more serious errors because the games couldn't even be playtested due to not running or crashing from any input. Overall, 38.9\% of all generated code was error-free, while 24.1\% was unplayable. However, as shown in the table, those rates are highly model-dependent.

\begin{table}[ht]
    \centering
    \caption{Distribution of perfectly implemented and unplayable games across models and code-generation modes. Rates are obtained by the sum of perfect or unplayable implementations over the total implementations per model, which is 36 for 12 games and 3 code-generation modes.}
    \begin{tabular}{c l c c c c c}
        \toprule
        \textbf{Model} & \textbf{Type} & \textbf{API} & \textbf{Indep.} & \textbf{Adap.} & \textbf{Sum/Total} & \textbf{Rate} \\
        \midrule
        \multirow{2}{*}{Claude} 
            & Perfect     & 5 & 9 & 6 & 20/36 & 55.6\% \\
            & Unplayable  & 0 & 0 & 0 & 0/36  & 0.0\%  \\
        \midrule
        \multirow{2}{*}{DeepSeek} 
            & Perfect     & 1 & 5 & 4 & 10/36 & 27.8\% \\
            & Unplayable  & 2 & 5 & 5 & 12/36 & 33.3\% \\
        \midrule
        \multirow{2}{*}{GPT} 
            & Perfect     & 4 & 5 & 3 & 12/36 & 33.3\% \\
            & Unplayable  & 4 & 2 & 8 & 14/36 & 38.9\% \\
        \bottomrule
    \end{tabular}
    \label{tab:success-failure}
\end{table}

\begin{figure}[ht]
    \centering
    \includegraphics[width=0.8\linewidth]{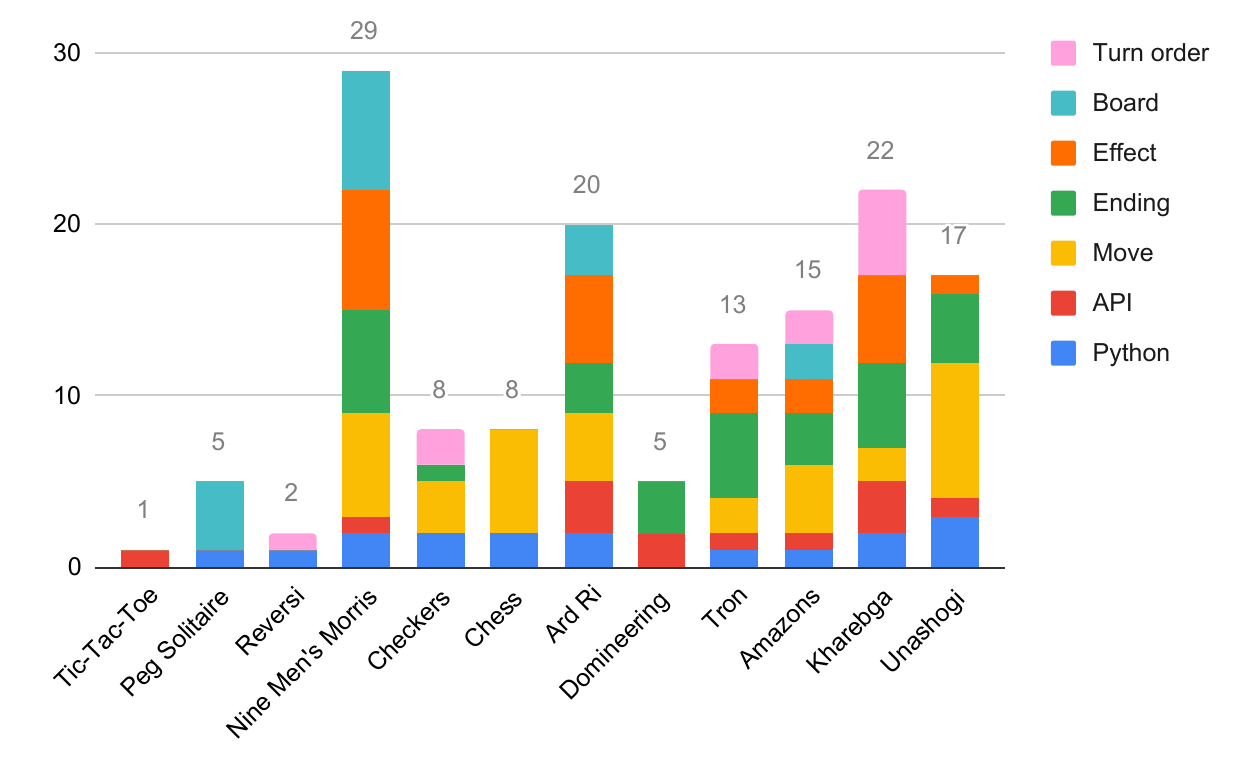}
    \caption{Occurrences of each error type per game, pooled from all experiments.}
    \label{fig:errors-per-game}
\end{figure}

Figure \ref{fig:errors-per-game} shows the sum total of errors for each game across all implementations. We can glean from it that Nine Men's Morris, Ard Ri, Kharebga and Unashogi were the hardest games for the models to implement, with these 4 games being responsible for 64\% of all errors. In fact, there were no perfect implementations of Nine Men's Morris or Unashogi.

Meanwhile, Tic-Tac-Toe, Peg Solitaire, Reversi and Domineering were the easiest to implement, and in fact account for 57\% of all perfect implementations. Claude implemented these four games perfectly in all experiments, with the same holding for GPT with Tic-Tac-Toe and Reversi. There was no such game for DeepSeek, but it provided the only case of an incorrect independent implementation being perfected in adaptation, with Checkers.

From these results, Claude had the best performance overall, with the fewest errors, the largest number of perfect implementations, and no game-breaking bugs. The other two models are approximately on par with each other.

Regarding our belief that the implementations of the popular games benefit from the models' existing knowledge of them, they account for 38.4\% of all errors, 55.8\% of all perfect implementations, and 28.0\% of all unplayable games. While that does show that the obscure games were harder to implement as a group, we also did not measure the intrinsic complexity of each game (how hard its logic is to represent in code), so a fair comparison would require further analysis.

\subsection{Qualitative Analysis} 
\label{sec:qualitative}
A common game-breaking error in implementations using the API was the models not defining the \texttt{next\_player} method. This is curious, as none of the other mandatory methods were ignored in any implementation. Furthermore, many adaptations kept references to methods or attributes defined only in the original implementation, which would also cause the game to crash. Some of these were among the easily fixable errors, as they would simply fail to reference an analogous existing method (e.g. calling \texttt{\_check\_game\_over} rather than \texttt{game\_finished}), and could be easily identified as the cause of the crash.

Another frequent error was separating each character on the board layout string with white spaces, which would conflict with the given board dimensions. These spaces were removed for test playing, with a Board error being noted for those implementations.

Both DeepSeek and GPT had a recurring problem in their implementations of Nine Men's Morris, Kharebga, and Unashogi: all of these games have piece placement in a phase, and piece capture as the goal in a separate phase. However, in many cases, the game would end as soon as the first move was made, as it simply counted that the opponent had no pieces without accounting for the first move.

Some errors occurred in most implementations of certain games by all models:
\begin{itemize}
    \item In Checkers, the forced capture rule was usually ignored.
    \item In Kharebga, multiple opposing pieces caught between the same pair of the player's pieces would incorrectly be captured all at once.
    \item In Unashogi, the Kinshō would be incorrectly able to move diagonally backwards.
\end{itemize}
All of these explicitly contradict the rules as described to the model in the prompts given, and the fact that at least one correct implementation of each exists proves that it is possible for the models to understand them properly. However, given the frequency of these errors, we acknowledge the possibility of the rule descriptions being insufficiently clear about these points.

Across all experiments, the most notably difficult game to implement was Nine Men's Morris. This is likely due to the graph-based nature of the game's board, which LLMs have trouble translating to a rectangular format, even despite efforts to facilitate that in the rules description. The overall trend was the inability to recognize two spaces as being adjacent, due to them being separated in the board layout for the sake of visibility.

While the API defines a standard format for inputting moves, it is also projected to be customizable in that regard. Therefore, deviations from the standard user interface were not counted as errors, so long as they did not impede usability. However, it is interesting to note that almost all of Claude's adapted implementations, save for Chess, kept the same interface used in the original implementation. The other models did not exhibit this behavior, with GPT always conforming the code to the standard input format, and DeepSeek doing the same to all but two games (Ard Ri and Amazons).

\section{Conclusion}
\label{sec:conclusion}
In this work, we introduced Boardwalk, a Python API for developing abstract board games, and examined the feasibility of using LLMs to generate board game code from natural language descriptions. We evaluated the implementation of 12 games by three LLMs in three code-generation modes: independent (free-form), API-compliant and adapting from free-form to API. 

 Overall, the answer to our research question (i.e., whether LLMs can implement digital versions of board games from rules described in natural language) is positive: Claude 3.7 Sonnet, the best-performing model, generated error-free code 55.6\% of the time. Furthermore, most erroneous results could be fixed with less effort than it would take to program the game from scratch, thus still providing a boost to efficiency. We find it feasible that, with additional work, this approach could lead to a functional and efficient framework for game code generation. %It is also very computationally efficient, requiring the model to be prompted only 2 to 4 times.

%Considering that the implementations evaluated were the immediate output of the models, without any modification, it should be possible to significantly improve these results with only a few iterations by giving the model simple feedback as to issues with the implementation. At its current stage, however, this workflow still represents an efficient time-saving option to programmers willing to playtest the results and fix the issues in the generated code.

While the tests involving the API generally showed a higher frequency of errors compared to the independent implementations, the usage of the API remains valid as a means to easily interface with both human and AI players, as well as a means to standardize and simplify the necessary code to develop a board game.

It is important to emphasize that these results were obtained from zero-shot inference experiments: no examples of code were provided to the models for context. In-context learning (giving several examples of the task at hand as part of the prompt) \cite{few-shot-learners} as well as few-shot learning (training the LLM for subsequent steps) greatly improve LLM performance. Thus, we intend to explore these approaches in future work, as we expect them to both increase the LLMs' understanding of the API and to better understand game descriptions in natural language. % results across the board, with minimal increase in computational complexity.

Another angle for future work to approach is to expand the API to allow for a larger breadth of representable games, such as those with hexagonal or graph-based boards, or card games. These could be potentially more challenging games and provide better insight into the LLMs' capacity for abstraction and reasoning. However, if a successful method to implement these is developed, it could serve as a reliable and versatile tool for implementing both existing and novel games, for researchers and game designers alike.

Future work could also investigate how our results could be improved with a few iterations, i.e., by giving the model feedback on its errors. This would be a step towards a fully-fledged AI board game creator. However, even at its current stage, the proposed approach represents an efficient assistant to programmers willing to create board games, as issues in the generated code can be manually fixed.

\section*{Acknowledgments}
The authors would like to acknowledge Eduardo Dalmas Faé for initial experiments on the topic. This study was financed in part by the Coordenação de Aperfeiçoamento de Pessoal de Nível Superior - Brasil (CAPES) - Finance Code 001 and the Conselho Nacional de Desenvolvimento Científico e Tecnológico (CNPq).

\bibliographystyle{sbc}
\bibliography{main}

\end{document}